\documentclass{bmvc2k}

\title{Group Relative Augmentation for Data Efficient Action Detection}

\addauthor{Deep Patel}{dpatel@nec-labs.com}{1}
\addauthor{Iain Melvin}{iain@nec-labs.com}{1}
\addauthor{Zachary Izzo}{zach@nec-labs.com}{1}
\addauthor{Martin Renqiang Min}{renqiang@nec-labs.com}{1}

\addinstitution{
NEC Laboratories America
}

\runninghead{Patel, Et al.}{Group Relative Augmentation}

\usepackage{booktabs} % For professional looking tables (\toprule, \midrule, \bottomrule)
\usepackage{siunitx}  % For aligning numbers by decimal points and handling units

\usepackage{comment}
\usepackage{amsmath}
\usepackage{amssymb}
\DeclareMathOperator{\BCEWithLogits}{BCEWithLogits}

\usepackage{booktabs}
\usepackage{xstring} 
\usepackage{etoolbox} % For \ifnumcomp

\usepackage{algorithm}
\usepackage[noend]{algpseudocode} % Use noend for conciseness

\begin{document}

\maketitle

\begin{abstract}
Adapting large Video-Language Models (VLMs) for action detection using only a few examples poses challenges like overfitting and the granularity mismatch between scene-level pre-training and required person-centric understanding. We propose an efficient adaptation strategy combining parameter-efficient tuning (LoRA) with a novel learnable internal feature augmentation. Applied within the frozen VLM backbone using FiLM, these augmentations generate diverse feature variations directly relevant to the task. Additionally, we introduce a group-weighted loss function that dynamically modulates the training contribution of each augmented sample based on its prediction divergence relative to the group average. This promotes robust learning by prioritizing informative yet reasonable augmentations. We demonstrate our method's effectiveness on complex multi-label, multi-person action detection datasets (AVA, MOMA), achieving strong mAP performance and showcasing significant data efficiency for adapting VLMs from limited examples.
\end{abstract}
\section{Introduction}
\label{sec:intro}

Large-scale pre-trained models, particularly Video-Language Models (VLMs), have demonstrated remarkable capabilities in understanding multimodal data, achieving state-of-the-art results on various video and language tasks \cite{wang2024internvideo2, cho2025perceptionlm, xu2021videoclip, yan2022videococa, bain2021frozen}. Leveraging these powerful models for downstream applications like spatio-temporal action detection is highly desirable, as it promises strong performance without requiring task-specific training from scratch on massive datasets. However, adapting these large models effectively, especially when labeled data for the target task is scarce, presents significant hurdles.

Our work focuses on the challenging yet practical scenario of adapting a pre-trained VLM for multi-label, multi-person action detection using only a few labeled examples per action class. This setting introduces several key difficulties. Firstly, fine-tuning the entire VLM, with its billions of parameters, on a small dataset inevitably leads to severe overfitting, hindering generalization. Parameter-efficient adaptation techniques are therefore essential. Secondly, many powerful VLMs are pre-trained on video text alignment tasks \cite{xu2021videoclip, yan2022videococa, luo2020univl, yang2023vid2seq}, which typically involve reasoning about the global scene content. In contrast, action detection demands fine-grained understanding, requiring the localization of specific individuals and the identification of their potentially concurrent actions within complex scenes involving multiple people. This represents a significant granularity mismatch between pre-training objectives and the downstream task.

Thirdly, while data augmentation is a standard technique to combat limited data, conventional augmentations applied at the input level (e.g., video transformations) may not be sufficient or optimally targeted for adapting the internal representations of a large frozen model. Generating diverse and task-relevant variations within the feature space could provide a more effective signal for adaptation, particularly for bridging the gap between scene-level pre-training and person-centric action detection.

To address these challenges, we propose an efficient VLM adaptation framework tailored for learning action detection from few examples. We employ Low-Rank Adaptation (LoRA) \cite{hu2022lora} for parameter-efficient fine-tuning, keeping the vast majority of the VLM frozen while only updating low-rank matrices injected into specific layers. Crucially, to enhance data diversity and guide adaptation effectively, we introduce a learnable feature augmentation module applied internally within the frozen vision encoder backbone. This module, based on Feature-wise Linear Modulation (FiLM) \cite{perez2018film}, learns to generate $N$ distinct augmentations of intermediate features, controlled by learnable embeddings.

Recognizing that not all generated augmentations may be equally beneficial, we further develop a novel training objective incorporating a \textit{group-weighted loss} mechanism. Instead of treating all augmented samples equally, this strategy dynamically calculates weights for each augmented sample based on the statistical divergence (e.g., using BCE distance) of its prediction compared to the average divergence across all augmentations for that specific instance. This allows the model to prioritize learning from augmentations that are informatively different yet statistically "reasonable" within the context of the learned transformations, while down-weighting potentially noisy or unhelpful ones. The objective also includes distillation and entropy terms to ensure consistency and promote diversity.

In summary, our main contributions are:
\begin{itemize}
    \item A parameter-efficient framework for adapting VLMs to multi-label, multi-person action detection using few examples, combining LoRA with learnable internal feature augmentation.
    \item A novel learnable feature augmentation module using FiLM that operates on intermediate VLM representations to generate diverse, task-relevant feature variations.
    \item A group-weighted loss strategy that intelligently utilizes augmented samples by dynamically weighting their contribution based on prediction divergence statistics within the augmentation group.
    \item Demonstration of the effectiveness of our approach on challenging, realistic action detection benchmarks (AVA \cite{gu2018ava}, MOMA \cite{luo2021moma}), showcasing improved data efficiency.
\end{itemize}

\section{Related Works}
\label{related_works}

\subsection{Efficient Adaptation of Vision Language Models}

Vision Foundation models pre-trained on extensive image-text pairs, such as CLIP \cite{radford2021learning} and ALIGN \cite{jia2021scaling}, and others \cite{zhai2023sigmoid, li2022blip, li2023blip, yu2022coca}, have revolutionized image understanding by demonstrating remarkable zero-shot and few-shot generalization capabilities across a wide array of downstream tasks. Inspired by this success, the video domain has also seen the emergence of strong Video-Language Models (VLMs) \cite{wang2024internvideo2, cho2025perceptionlm, xu2021videoclip, yan2022videococa}. These models are typically trained on large-scale datasets of video-text pairs, often incorporating self-supervised learning objectives such as masked autoencoding \cite{tong2022videomae} to learn robust spatio-temporal representations.

Given the significant computational resources required for pre-training these large VLMs, efficiently adapting them to specific downstream tasks with limited labeled data has become a critical area of research. Several approaches have been proposed for adapting image and video-based VLMs to specific domains and downstream tasks \cite{chen2022visualgpt, zhu2023not, wu2023feature, zhang2021tip, gao2024clip, iCLIP_ICCVW23}. The success of vision-language models in open-vocabulary object detection \cite{kim2023region, zareian2021open, wu2023cora, lin2022learning, zhong2022regionclip} has spurred interest in leveraging VLMs for more granular video understanding tasks. Some efforts have focused on adapting VLMs for zero-shot action detection \cite{bao2025exploiting, huang2023interaction, huang2025spatio}, where the model identifies actions without seeing any examples of those specific actions during fine-tuning, relying instead on learned associations between visual content and textual descriptions of actions.

However, adapting large VLMs for multi-label, multi-person action detection from few examples remains largely unexplored. This task demands fine-grained, instance-level understanding and localization of multiple actors and assigning them potentially concurrent actions which often exceeds the capabilities of models pre-trained for scene-level understanding.

\subsection{Data and Feature Augmentation as Regularizers}

Data augmentation is a well-established technique to improve model generalization in vision tasks, ranging from simple input transformations like cropping and color jittering \cite{yang2022image} to more advanced methods such as Mixup, CutMix, and AutoAugment \cite{yun2019cutmix, cubuk2019autoaugment}. These methods primarily modify input data to create diverse training examples.

Recently, augmenting representations directly within the model's feature space has emerged as a powerful regularization strategy, especially for adapting pre-trained models with limited data. Early work in this area includes adversarial training, which adds perturbations to features to enhance robustness.  Approaches, like A-FAN \cite{chen2021adversarial} and FASA \cite{zang2021fasa}, have explored adversarial feature perturbations, sometimes guided by feature statistics or normalization techniques, for tasks such as visual recognition and instance segmentation. FroFA \cite{bar2024frozen} demonstrated that simple pointwise transformations applied to frozen image features can improve few-shot classification. Some methods also leverage external semantics, like TextManiA \cite{ye2023textmania} using language models to guide visual feature augmentation, or Moment Exchange \cite{li2021feature} manipulating feature statistics to encourage learning from style and shape cues.

While these feature augmentation techniques are insightful, most have focused on the image domain. Extending these to video requires addressing temporal consistency. For instance, Motion Coherent Augmentation (MCA) \cite{zhang2024don} aims to modify video appearances while preserving motion information. Our work builds on the idea of internal feature augmentation but tailors it for adapting large Video-Language Models to action detection from few examples. Unlike prior methods relying on predefined transformations or purely adversarial noise in the feature space, our augmentations are learned and controlled by distinct embeddings, and their impact is managed by a novel group-weighted loss. This allows for task-relevant feature variations directly within the VLM, specifically for adapting to person-centric actions.

\section{Method}
\label{sec:method}

We propose a parameter-efficient method for adapting large pre-trained Video-Language Models (VLMs) to downstream action detection tasks using only a few examples. Our approach leverages Low-Rank Adaptation (LoRA) for efficient fine-tuning, introduces learnable feature augmentations applied internally within the vision encoder, and employs a novel group augmentation weighting strategy within the training objective to enhance robustness and generalization.

%-------------------------------------------------------------------------
\subsection{Model Architecture and Adaptation}
\label{sec:model_adaptation}

We utilize a pre-trained VLM, keeping its parameters largely frozen, and focus adaptation on the vision encoder component. To mitigate overfitting and reduce computational cost in the few-shot setting, we employ Low-Rank Adaptation (LoRA)~\cite{hu2022lora}.

LoRA injects trainable low-rank matrices into specific layers. For a pre-trained weight matrix $W_0 \in \mathbb{R}^{d \times k}$, the adapted weight $W$ is parameterized as $W = W_0 + BA$, where $B \in \mathbb{R}^{d \times r}$ and $A \in \mathbb{R}^{r \times k}$ are trainable low-rank matrices with rank $r \ll \min(d, k)$, while $W_0$ remains frozen.

We apply LoRA to the query ($W_q$) and value ($W_v$) projection matrices within the multi-head self-attention (MHSA) modules of the vision encoder's transformer blocks, specifically from layer index $l_{lora}$ onwards. This focuses adaptation on higher-level visual features. Let $\theta$ denote the frozen pre-trained parameters of the VLM and $\phi_{lora}$ represent the set of all trainable LoRA parameters (the matrices $A$ and $B$ across the adapted layers). The overall adapted model computes outputs based on both $\theta$ and $\phi_{lora}$.

%-------------------------------------------------------------------------
\subsection{Learnable Feature Augmentation}
\label{sec:feature_augmentation}

To enhance generalization from limited data, we introduce augmentations directly in the feature space of the vision encoder. We insert a learnable feature augmentation module based on Feature-wise Linear Modulation (FiLM)~\cite{perez2018film} after an intermediate layer $l_{aug}$ of the frozen encoder backbone.

Let $f^{(l_{aug})} \in \mathbb{R}^{B \times T' \times C}$ be the output feature map of layer $l_{aug}$ for a batch of $B$ input sequences, where $T'$ is the sequence length of visual tokens and $C$ is the feature dimension. The augmentation module generates $N$ distinct augmented versions of these features. It comprises:
\begin{enumerate}
    \item A learnable embedding table $E \in \mathbb{R}^{N \times D_e}$, where $N$ is the number of augmentation types and $D_e$ is the embedding dimension. Each row $e_i$ represents the $i$-th augmentation embedding.
    \item A FiLM parameter generator $g_\psi: \mathbb{R}^{D_e} \to \mathbb{R}^{2C}$, implemented as a small trainable Multi-Layer Perceptron (MLP) with parameters $\psi$. For each embedding $e_i$, it predicts modulation parameters $(\gamma_i, \beta_i)$:
    \begin{equation}
        (\gamma_i, \beta_i) = g_\psi(e_i), \quad \gamma_i, \beta_i \in \mathbb{R}^{C}
    \end{equation}
    The parameters $\psi$ and the embedding table $E$ constitute the trainable parameters $\phi_{aug}$ of this module. The generator $g_\psi$ is initialized such that $\gamma_i \approx \mathbf{1}$ and $\beta_i \approx \mathbf{0}$ initially for all $i=1, \dots, N$.
    \item A FiLM application step where each $(\gamma_i, \beta_i)$ modulates the original features $f^{(l_{aug})}$ to produce the $i$-th augmented feature map $f^{(l_{aug})}_{\text{aug}, i}$:
    \begin{equation}
        f^{(l_{aug})}_{\text{aug}, i} = \gamma_i \odot f^{(l_{aug})} + \beta_i
        \label{eq:film}
    \end{equation}
    where $\odot$ denotes element-wise multiplication with broadcasting over the batch and sequence dimensions.
\end{enumerate}

During training, the original features $f^{(l_{aug})}$ and all $N$ augmented feature maps $\{f^{(l_{aug})}_{\text{aug}, i}\}_{i=1}^N$ are passed through the subsequent layers of the network (which include the trainable LoRA parameters $\phi_{lora}$). This yields output logits $z_{\text{orig}} \in \mathbb{R}^{B \times K}$ corresponding to the original features and a set of augmented logits $\{z_{\text{aug}, i} \in \mathbb{R}^{B \times K}\}_{i=1}^N$, where $K$ is the number of output classes. The trainable parameters during optimization are $\phi = \phi_{lora} \cup \phi_{aug}$.

%-------------------------------------------------------------------------
\subsection{Training Objective}
\label{sec:loss}

Our training objective combines standard supervision with regularization terms derived from the augmented samples to promote robustness and diversity. The total loss $\mathcal{L}_{\text{total}}$ is a weighted sum of four components:

\begin{equation}
    \mathcal{L}_{\text{total}} = \mathcal{L}_{\text{BCE}} + \lambda_{\text{distill}} \mathcal{L}_{\text{distill}} + \lambda_{\text{aug}} \mathcal{L}_{\text{BCE-W}} + \lambda_{\text{ent}} \mathcal{L}_{\text{ent}}
    \label{eq:total_loss}
\end{equation}
where $\lambda_{\text{distill}}, \lambda_{\text{aug}}, \lambda_{\text{ent}}$ are scalar hyper-parameters. Let $y \in \{0, 1\}^{B \times K}$ be the ground truth multi-label matrix for the batch.

\paragraph{1. Anchor Task Loss ($\mathcal{L}_{\text{BCE}}$):} The standard supervised loss applied to the predictions from the original, unaugmented features $z_{\text{orig}}$. For multi-label classification, we use the Binary Cross-Entropy (BCE) loss with logits:
\begin{equation}
    \mathcal{L}_{\text{BCE}} = \frac{1}{B K} \sum_{b=1}^B \sum_{k=1}^K \BCEWithLogits(z_{\text{orig}, b, k}, y_{b,k})
    \label{eq:loss_bce_orig}
\end{equation}
where $z_{\text{orig}, b, k}$ is the logit for the $k$-th class of the $b$-th sample.

\paragraph{2. Distillation Loss ($\mathcal{L}_{\text{distill}}$):} A regularization term encouraging predictions from augmented features ($z_{\text{aug}, i}$) to remain consistent with those from the original features ($z_{\text{orig}}$). We use the sigmoid probabilities derived from $z_{\text{orig}}$ as soft targets, detaching them from the gradient computation. Let $p_{\text{orig}, b} = \sigma(z_{\text{orig}, b})_{\text{detached}}$, where $\sigma(\cdot)$ denotes the element-wise sigmoid function. The loss is the average BCE between augmented logits and these soft targets:
\begin{equation}
    \mathcal{L}_{\text{distill}} = \frac{1}{N B K} \sum_{i=1}^N \sum_{b=1}^B \sum_{k=1}^K \BCEWithLogits(z_{\text{aug}, i, b, k}, p_{\text{orig}, b, k}).
    \label{eq:loss_distill}
\end{equation}

\paragraph{3. Weighted Augmented Loss ($\mathcal{L}_{\text{BCE-W}}$):} The supervised BCE loss applied to predictions from augmented features, modulated by a per-sample weight $w_{i,b} \in [0, 1]$ (derived in Sec.~\ref{sec:weighting}) reflecting the "reasonableness" of the $i$-th augmentation for the $b$-th sample. Let $\ell_{i,b}$ be the average BCE loss for the augmented sample $(i, b)$:
\begin{equation}
    \ell_{i,b} = \frac{1}{K} \sum_{k=1}^K \BCEWithLogits(z_{\text{aug}, i, b, k}, y_{b,k})
    \label{eq:bce_aug_per_sample}
\end{equation}
The total weighted augmented loss is:
\begin{equation}
    \mathcal{L}_{\text{BCE-W}} = \frac{1}{N B} \sum_{i=1}^N \sum_{b=1}^B w_{i,b} \, \ell_{i,b}
    \label{eq:loss_bce_weighted}
\end{equation}

\paragraph{4. Entropy Maximization Loss ($\mathcal{L}_{\text{ent}}$):} A loss term encouraging diversity and uncertainty in the predictions derived from augmented features. We minimize the negative average binary entropy of the predicted probabilities for augmented samples:
\begin{equation}
    \mathcal{L}_{\text{ent}} = - \frac{1}{N B K} \sum_{i=1}^N \sum_{b=1}^B \sum_{k=1}^K H(\sigma(z_{\text{aug}, i, b, k}))
    \label{eq:loss_entropy}
\end{equation}
where $H(p) = -[p \log p + (1-p) \log(1-p)]$ is the binary entropy function.

%-------------------------------------------------------------------------
\subsection{Group Augmentation Weighting}
\label{sec:weighting}

The weights $w_{i,b}$ in Eq.~\ref{eq:loss_bce_weighted} dynamically modulate the contribution of each augmented sample based on its predictive divergence from the original sample, relative to other augmentations in the group. We first measure the dissimilarity between the $i$-th augmented prediction and the (detached) original prediction for sample $b$ using the average BCE distance $d_{i,b}$:
\begin{equation}
    d_{i,b} = \frac{1}{K} \sum_{k=1}^K \BCEWithLogits(z_{\text{aug}, i, b, k}, p_{\text{orig}, b, k})
    \label{eq:bce_distance}
\end{equation}
where $p_{\text{orig}, b} = \sigma(z_{\text{orig}, b})_{\text{detached}}$.

For each sample $b$, we compute the mean $\mu_{d,b}$ and standard deviation $\sigma_{d,b}$ of these distances across the $N$ augmentations:
\begin{align}
    \mu_{d,b} &= \frac{1}{N} \sum_{i=1}^N d_{i,b} \\
    \sigma_{d,b} &= \sqrt{\frac{1}{N} \sum_{i=1}^N (d_{i,b} - \mu_{d,b})^2}
\end{align}
The weight $w_{i,b}$ is calculated using a Gaussian function centered at the mean distance $\mu_{d,b}$, based on the z-score of the distance $d_{i,b}$:
\begin{equation}
    z_{i,b} = \frac{d_{i,b} - \mu_{d,b}}{\sigma_{d,b} + \epsilon}
\end{equation}
\begin{equation}
    w_{i,b} = \exp\left(-\frac{z_{i,b}^2}{2 s^2}\right)
    \label{eq:weight_calc}
\end{equation}
where $s$ is a hyper-parameter controlling the sensitivity to deviations from the mean distance, and $\epsilon$ is a small constant (e.g., $10^{-6}$) for numerical stability. This strategy assigns higher weights to augmentations whose predictive divergence is close to the average divergence observed for that sample across all $N$ augmentations.

%-------------------------------------------------------------------------
\begin{comment}
\subsection{Implementation Details}
\label{sec:implementation}

The VLM backbone is initialized with pre-trained weights. The trainable LoRA matrices $A$ and $B$ are initialized such that the initial LoRA update $\Delta W = BA$ is zero. The FiLM generator $g_\psi$ is initialized to produce near-identity transformations ($\gamma_i \approx \mathbf{1}, \beta_i \approx \mathbf{0}$). Key hyper-parameters include the number of learned augmentations $N=10$, the augmentation embedding dimension $D_e=128$, the vision encoder layer index for feature augmentation $l_{aug}=30$, and the starting layer index for LoRA adaptation $l_{lora}$. The loss weights are set to $\lambda_{\text{distill}}=0.1$, $\lambda_{\text{aug}}=1.0$, and $\lambda_{\text{ent}}=0.01$. The weighting function scale parameter is $s=1.5$. We use the AdamW optimizer~\cite{??} to train the parameters $\phi = \phi_{lora} \cup \phi_{aug}$. A temperature parameter potentially used within the model architecture can be annealed during training, for instance, following a cosine decay schedule.
    
\end{comment}

\section{Experiments}
\label{results}

\subsection{Datasets and Metrics}
\label{sec:datasets_metrics}

To evaluate the performance of our method, we focus on challenging benchmarks that reflect realistic scenarios involving multiple people performing potentially multiple actions simultaneously. We utilize two primary datasets:

\begin{itemize}
    \item \textbf{AVA (Atomic Visual Actions) v2.2~\cite{gu2018ava}}: AVA is a large-scale dataset specifically designed for spatio-temporal action detection. It features densely annotated keyframes from movies, with bounding boxes localizing individuals and multi-label annotations of 80 atomic visual actions for each person within the box.
    
    \item \textbf{MOMA (Multi-Object Multi-Actor) 1.0~\cite{luo2021moma}}: offers rich annotations for human activities, including sub-activities, and atomic actions, along with instance-level actor and object details. For our experiments, we adapt MOMA to align with the AVA dataset's format by focusing on person-centric action labels at the keyframe level. This results in a dataset suitable for multi-label action detection with 52 distinct action classes.

\end{itemize}

\paragraph{Experimental Setup and Evaluation}
We evaluate our proposed method for adapting the InternVideo2 (40-layer) video encoder \cite{wang2024internvideo2} on the AVA and MOMA action detection datasets. The adaptation is performed in a data-efficient setting, using approximately $K=15$ training examples sampled per action class from the official training splits. The text encoder remains frozen throughout all experiments. \footnote{More information about sampling training data is available in the supplementary material}

We compare our approach against two methods: (1) a baseline using LoRA with standard input data augmentations, and (2) a recent video augmentation technique, MCA \cite{zhang2024don}. Both comparison methods apply LoRA to all vision encoder layers, resulting in 3.60M trainable parameters. In contrast, our method strategically applies LoRA to only the latter half of the vision encoder (from layer 30 onwards), significantly reducing trainable parameters to 1.27M.\footnote{For the LoRA + Std. Aug. baseline and MCA, applying LoRA to all vision encoder layers yielded better performance than applying it only to the latter half, unlike our method which benefits from this targeted application.} Performance is evaluated using the standard mean Average Precision (mAP) metric.

\subsection{Results}
\label{sec:results}

% Table 1: AVA Results
\begin{table}[t] % [t] suggests placing the table at the top of the page
    \centering
    \sisetup{table-format=2.2, table-space-text-post = \si{\percent}} % Setup for mAP column alignment
    \begin{tabular}{@{}l c c S[table-format=1.2, table-space-text-post=m] S@{}} % Define columns
        \toprule
        Method                      & {Params (M)} & {mAP (\%)} \\ % Column headers
        \midrule
        LoRA + Std. Aug.            & 3.60m          & 24.30      \\ % Use {} for non-numeric S column entry
        LoRA + Ours (FiLM, N=10)   & 1.27m        & 25.56      \\
        LoRA + MCA                 & 3.60m        & 25.63      \\
        \bottomrule
    \end{tabular}
    \caption{Comparison of adaptation methods on the AVA dataset. All experiments use 15 examples per class and 8 input frames.}
    \label{tab:ava_results}

\end{table}

As shown in Table~\ref{tab:ava_results}, on the AVA dataset, our method achieves an mAP of 25.56\%, significantly outperforming the LoRA + Std. Aug. baseline (24.30\%). This represents a +1.26\% absolute improvement. Notably, our approach performs comparably to LoRA + MCA (25.63\%) while using approximately one-third of the trainable parameters (1.27M vs 3.60M).

% Table 2: MOMA Results
\begin{table}[t]
    \centering
    \sisetup{table-format=2.1, table-space-text-post = \si{\percent}} % Setup for mAP column alignment
    \begin{tabular}{@{}l c c S[table-format=1.2, table-space-text-post=m] S@{}} % Define columns
        \toprule
        Method                    & {Params (M)} & {mAP (\%)} \\ % Column headers
        \midrule
        LoRA + Std. Aug.          & 3.60m        & 23.7       \\
        LoRA + Ours (FiLM, N=8)   & 1.27m        & 24.6       \\
        LoRA + MCA                & 3.60m        & 24.3       \\
        \bottomrule
    \end{tabular}
    \caption{Comparison of adaptation methods on the MOMA dataset. All experiments use 15 examples per class and 8 input frames.}
    \label{tab:moma_results}
\end{table}

On the MOMA dataset (Table~\ref{tab:moma_results}), our method with $N=8$ learned augmentations similarly improves upon the baseline. It scores 24.6\% mAP, a +0.9\% improvement over the LoRA + Std. Aug. baseline (23.7\%). Furthermore, our method surpasses LoRA + MCA (24.3\%) by +0.3\% mAP, again with substantially fewer learnable parameters. These results demonstrate the effectiveness and parameter efficiency of our proposed internal feature augmentation and group weighting strategy for adapting VLMs from few examples.

\subsection{Ablations}
\label{sec:ablations}

% Table 3: Ablation Study
\begin{table}[t]
    \centering
    \sisetup{table-format=2.1, table-space-text-post = \si{\percent}} % Setup for mAP column alignment
    \begin{tabular}{@{}c S@{}} % Define columns - centered number, S column for mAP
        \toprule
        \# Augmentations ($N$) & {mAP (\%)} \\ % Column headers
        \midrule
        2                    & 24.1       \\
        5                    & 24.1       \\
        8                    & 24.6       \\
        10                   & 24.5       \\
        12                   & 24.3       \\
        \bottomrule
    \end{tabular}
    \caption{Ablation study on the impact of the number of learned augmentations ($N$) on mAP (\%), evaluated on the MOMA dataset.}   \label{tab:ablation_num_aug}
\end{table}

We conduct ablation studies to validate the design choices of our method.

\paragraph{Impact of Number of Learned Augmentations ($N$).}
Table~\ref{tab:ablation_num_aug} shows the mAP on MOMA when varying the number of learned augmentations ($N$). The performance peaks at $N=8$, achieving 24.6\% mAP. Using fewer augmentations (e.g., $N=2$ or $N=5$) results in a lower mAP of 24.1\%. Increasing $N$ beyond 8 to 10 or 12 slightly degrades performance (24.5\% and 24.3\% respectively), suggesting that a moderate number of diverse learned augmentations is optimal, and too many might introduce noise or redundancy.

\begin{table}[t]
    \centering
    \begin{tabular}{@{}l S S[table-format=2.2, explicit-sign, print-zero-exponent=true, table-sign-mantissa, table-space-text-post = \si{\percent}]@{}}
        \toprule
        Method Component(s)                      & {mAP (\%)} \\
        \midrule
        Baseline (LoRA + Std. Aug.)                                & 24.30  \\

        + $\mathcal{L}_{\text{BCE}}$ (w/o group weights)           & 24.97           \\
        + $\mathcal{L}_{\text{BCE-W}}$ (with group weights)         & 24.96         \\
        \addlinespace
        + $\mathcal{L}_{\text{distill}}$ only                       & 25.23        \\
        + $\mathcal{L}_{\text{ent}}$ only                          & 25.36       \\
        \addlinespace
        + $\mathcal{L}_{\text{distill}} + \mathcal{L}_{\text{ent}} + \mathcal{L}_{\text{BCE}}$ (w/o group weights) & 25.26 \\
        + $\mathcal{L}_{\text{distill}} + \mathcal{L}_{\text{ent}} + \mathcal{L}_{\text{BCE-W}}$ (with group weights) & 25.56 \\
        \bottomrule
    \end{tabular}
    \caption{Ablation study on the impact of different loss components on mAP (\%), evaluated on the AVA dataset.}
    \label{tab:ablation_loss_components}
\end{table}

\paragraph{Impact of Loss Components.}
We ablate the individual and combined effects of our proposed loss terms on the AVA dataset, with results shown in Table~\ref{tab:ablation_loss_components}. The baseline (LoRA + Std. Aug.) achieves 24.30\% mAP.
Adding only the BCE loss for augmented samples ($\mathcal{L}_{\text{BCE}}$), without our group weighting, improves performance to 24.97\%. Introducing group weighting to this term ($\mathcal{L}_{\text{BCE-W}}$) yields a similar 24.96\%.
Individually, the distillation loss ($\mathcal{L}_{\text{distill}}$) and the entropy loss ($\mathcal{L}_{\text{ent}}$) provide substantial gains, reaching 25.23\% (+0.93\%) and 25.36\% (+1.06\%) mAP respectively, highlighting their importance in regularizing augmentations and promoting diversity.
Combining all components, the full model with $\mathcal{L}_{\text{distill}} + \mathcal{L}_{\text{ent}} + \mathcal{L}_{\text{BCE-W}}$ (with group weights) achieves the best performance of 25.56\% (+1.26\% over baseline). 

\begin{comment}
This demonstrates that all proposed loss components contribute positively to the final performance, with the group weighting providing a crucial element when all losses are combined to effectively manage the impact of augmented samples. The version without group weights in the full combination also improves (25.26\%), but the group weighting adds an additional +0.3\% mAP, confirming its utility.
\end{comment}

\section{Conclusion \& Limitations}

In this work, we addressed the challenge of adapting large pre-trained Video-Language Models for complex multi-label, multi-person action detection using only a few examples per class. We proposed an efficient adaptation strategy that combines parameter-efficient fine-tuning using LoRA with learnable internal feature augmentation mechanism. This FiLM-based augmentation, applied to intermediate features of the frozen VLM backbone, generates diverse task-relevant variations. To effectively leverage these, we introduced a group-weighted loss function, alongside distillation and entropy regularization, which dynamically modulates the contribution of augmented samples based on their predictive consistency.

Our experiments on the challenging AVA and MOMA datasets demonstrate that our approach significantly outperforms standard LoRA with conventional data augmentation. Notably, our method achieves performance comparable to SOTA video augmentation methods like MCA while utilizing less trainable parameters, highlighting its parameter efficiency.

Despite these promising results, our work has limitations. Firstly, optimal hyper-parameters, such as the number of learned augmentations ($N$) and the loss component weights ($\lambda$), are dataset-dependent and require empirical tuning. Secondly, while effective for action detection, the generalization of our learned augmentation and weighting strategies to other VLM architectures or diverse downstream video tasks needs further investigation.

\bibliography{bmvc_review}

\title{Supplementary Material}

\maketitle

\section{Multi Label Training Data Sampling}
We evaluate our model in a data efficient learning setting. For the training (support) set, we aim to sample approximately $K=15$ examples per action class from the official training splits of MOMA and AVA. However, due to the multi-label nature of both datasets, where a single person instance can be associated with multiple action labels, strictly guaranteeing exactly $K$ examples per class is challenging. Our sampling strategy iteratively collects person instances associated with each action class. Once the target count for a specific class is reached, we stop adding new instances primarily for that class, though instances added for other under-sampled classes might still carry labels for already sufficiently sampled classes. This results in a support set where most classes have approximately $K$ examples, but the exact counts may vary slightly depending on label co-occurrence.%
\end{document}